\documentclass[10pt,twocolumn,letterpaper]{article}

\usepackage[pagenumbers]{cvpr}

\usepackage{graphicx}
\usepackage{amsmath}
\usepackage{amssymb}
\usepackage{booktabs}
\usepackage{multirow}
\usepackage{algorithm}
\usepackage{algorithmic}
\usepackage{xcolor}
\usepackage{enumitem}
\usepackage[breaklinks=true,bookmarks=false]{hyperref}

\def\cvprPaperID{****}
\def\confName{CVPR}
\def\confYear{2025}

\begin{document}

\title{Multi-Branch Non-Homogeneous Image Dehazing via \\
Concentration Partitioning and Image Fusion}

\author{
Yingming Zhang$^{1}$ \quad Wuqi Su$^{1}$ \quad Qing Xiao$^{2*}$ \quad Yonggang Yang$^{2}$ 
\\
$^{1}$School of Software and Internet of Things Engineering, Zhejiang Gongshang University \\
$^{3}$School of Computer Science and Technology, Tiangong University \\
{\tt\small \{zhangyingming, suwuqi\}@zjgsu.edu.cn}
}

\maketitle

\begin{abstract}
Existing single image dehazing methods have demonstrated satisfactory performance on homogeneous thin-haze images; however, they often struggle with non-homogeneous hazy images that exhibit spatially varying haze concentrations and abrupt density transitions across different regions. To address this fundamental limitation, we propose a novel multi-branch deep neural network framework, termed Concentration Partitioning and Image Fusion Network (CPIFNet), which decomposes the challenging non-homogeneous dehazing problem into a set of tractable homogeneous sub-problems. Our key insight is that a single non-homogeneous hazy image can be viewed as a composite of multiple local regions, each exhibiting approximately homogeneous haze characteristics. CPIFNet employs a two-stage architecture consisting of an Image Enhancement Network (IENet) stage and an Image Fusion Network (IFNet) stage. In the first stage, multiple IENet branches are independently trained on homogeneous haze datasets of different concentration levels, producing enhancement models that excel at restoring regions matching their respective haze densities. In the second stage, the IFNet intelligently aggregates the advantageous regions from all enhancement outputs through deep feature stacking and merging, yielding a unified high-quality dehazed result. Furthermore, we introduce a comprehensive loss function incorporating reconstruction, perceptual, structural, and color losses to jointly supervise both stages. Extensive experiments on both synthetic benchmarks (FiveK-Haze, SOTS-indoor, SOTS-outdoor) and real-world datasets demonstrate that CPIFNet achieves state-of-the-art performance, surpassing the second-best method by 5.29 dB in PSNR on FiveK-Haze, 2.52 dB on SOTS-indoor, and 2.83 dB on SOTS-outdoor, while producing visually superior results on real-world non-homogeneous hazy images with enhanced structural contrast and natural color reproduction.
\end{abstract}

\section{Introduction}
\label{sec:intro}

Outdoor images captured under adverse weather conditions are frequently degraded by atmospheric particles such as smoke, dust, and haze, leading to significant deterioration in image clarity, color fidelity, and structural contrast~\cite{narasimhan2000chromatic,he2009single}. Such degraded images not only hinder human visual perception but also adversely impact downstream high-level computer vision tasks, including autonomous driving~\cite{li2019benchmarking}, remote sensing monitoring~\cite{liang2024remote}, scene text detection~\cite{tang2022few,tang2022you}, document image understanding~\cite{feng2023docpedia,tang2024textsquare,zhao2024tabpedia}, and outdoor surveillance anomaly detection. Consequently, image dehazing has emerged as a critical preprocessing step for numerous real-world visual systems. Early physics-based approaches primarily relied on the atmospheric scattering model (ASM)~\cite{narasimhan2000chromatic}, which formulates the haze degradation process as:
\begin{equation}
I(p) = J(p) \times t(p) + A(1 - t(p)),
\label{eq:asm}
\end{equation}
where $p$ denotes the pixel location, $I$ and $J$ represent the hazy and haze-free images respectively, $A$ is the global atmospheric light, and $t$ is the transmission map. Based on this model, He~\emph{et al.}~\cite{he2009single,he2011single} proposed the seminal dark channel prior (DCP), which exploits the statistical observation that most local patches in haze-free outdoor images contain pixels with very low intensities in at least one color channel. Following this pioneering work, Zhu~\emph{et al.}~\cite{zhu2015fast} introduced the color attenuation prior for rapid single image dehazing, while Zhao~\emph{et al.}~\cite{zhao2016dark} developed an improved algorithm incorporating atmospheric light validation and halo elimination. Shen~\emph{et al.}~\cite{shen2017atmospheric} proposed atmospheric light correction and transmission optimization to address the sky darkening artifact in dehazed images, and their subsequent work~\cite{shen2017integrating} integrated sky detection with texture smoothing for further improvement. Cao~\emph{et al.}~\cite{cao2018fast} presented a fast dehazing algorithm based on luminance contrast enhancement and saturation compensation. However, these handcrafted prior-based methods are inherently limited by human observation and fail to generalize across diverse real-world scenarios.

The advent of deep learning has fundamentally transformed the image dehazing landscape. DehazeNet~\cite{cai2016dehazenet} pioneered the use of convolutional neural networks (CNNs) for learning the medium transmission map in an end-to-end manner. Ren~\emph{et al.}~\cite{ren2016single} proposed a multi-scale CNN architecture that leverages coarse-to-fine processing for improved haze removal. Li~\emph{et al.}~\cite{li2017aod} reformulated the atmospheric scattering model and proposed AOD-Net, which directly predicts the mapping coefficient between hazy and clean images. Qu~\emph{et al.}~\cite{qu2019enhanced} developed an enhanced pix2pix dehazing network leveraging generative adversarial learning~\cite{goodfellow2014generative}. Qin~\emph{et al.}~\cite{qin2020ffa} designed the Feature Fusion Attention Network (FFA-Net), which stacks extracted features to directly enhance image contrast. With the rise of attention mechanisms~\cite{vaswani2017attention} and vision transformers~\cite{dosovitskiy2021image,liu2021swin}, transformer-based approaches have also been explored for dehazing. Song~\emph{et al.}~\cite{song2023deformer} proposed DeFormer, a vision transformer tailored for single image dehazing, achieving competitive results on benchmark datasets. Zamir~\emph{et al.}~\cite{zamir2022restormer} introduced Restormer for efficient high-resolution image restoration including dehazing, and Liang~\emph{et al.}~\cite{liang2021swinir} proposed SwinIR exploiting the hierarchical architecture of Swin Transformer~\cite{liu2021swin}. Multi-modal learning and visual understanding methods~\cite{tang2025mtvqa,zhao2023multi,tang2023character,lu2025bounding,feng2025dolphin} have also inspired new perspectives on feature extraction and fusion strategies in low-level vision.

Despite these remarkable advances, existing methods predominantly treat the entire image uniformly and are primarily designed and evaluated on homogeneous haze scenarios. In real-world environments, however, haze is inherently non-homogeneous---the haze concentration varies dramatically across different spatial locations within a single image, with some regions exhibiting thick dense haze while others contain only light mist or are nearly haze-free. This spatial non-homogeneity poses a fundamental challenge: a model tuned for dense haze tends to over-enhance light haze regions, while a model optimized for light haze provides insufficient restoration for densely hazed areas. Several recent efforts have attempted to address non-homogeneous dehazing. Zhang~\emph{et al.}~\cite{zhang2022hazdesnet} designed a two-branch network where one branch estimates the haze density map to guide the other branch for dehazing. Zheng~\emph{et al.}~\cite{zheng2023curricular} established a physics-aware dual-branch unit based on the atmospheric scattering model to improve non-homogeneous dehazing. Liang~\emph{et al.}~\cite{liang2024remote} proposed a superpixel-based atmospheric light fusion estimation model for more accurate atmospheric light estimation. Guo~\emph{et al.}~\cite{guo2023scanet} leveraged brightness discrepancies to constrain attention maps for high-quality reconstruction of densely hazed regions. Chen~\emph{et al.}~\cite{chen2024dea} proposed a detail-enhanced attention block composed of detail-enhanced convolution and content-guided attention for improved dehazing. However, these methods still process the entire image through a single network pathway, which limits their ability to simultaneously handle regions of vastly different haze concentrations.

In this paper, we propose a fundamentally different approach to non-homogeneous image dehazing. Our key insight is that \textbf{a single non-homogeneous hazy image can be decomposed into multiple local regions, each exhibiting approximately homogeneous haze characteristics}. Based on this observation, we design CPIFNet, a two-stage, multi-branch neural network framework that decomposes the non-homogeneous dehazing problem into a set of homogeneous sub-problems. The main contributions of this work are summarized as follows:

\begin{itemize}[leftmargin=*,nosep]
\item We propose a novel decomposition strategy that treats a single non-homogeneous hazy image as a composite of multiple homogeneous hazy regions, enabling targeted restoration for each concentration level.
\item We design the Image Enhancement Network (IENet), which is independently trained on homogeneous haze datasets of different concentration levels, producing models with varying enhancement intensities tailored to specific haze densities.
\item We design the Image Fusion Network (IFNet), which intelligently aggregates the advantageous regions from multiple IENet enhancement outputs through deep feature stacking and merging, producing a unified high-quality dehazed result.
\item Comprehensive experiments on both synthetic and real-world benchmarks demonstrate that CPIFNet significantly outperforms state-of-the-art methods, achieving substantial improvements across all evaluation metrics.
\end{itemize}

\section{Related Work}
\label{sec:related}

\subsection{Prior-Based Image Dehazing}

Traditional image dehazing methods primarily rely on handcrafted priors derived from statistical observations of natural haze-free images. He~\emph{et al.}~\cite{he2009single,he2011single} proposed the influential dark channel prior (DCP), which observes that in most non-sky local patches of haze-free outdoor images, at least one color channel has very low pixel intensity. This prior enables estimation of the transmission map and atmospheric light from a single hazy image. Zhu~\emph{et al.}~\cite{zhu2015fast} introduced the color attenuation prior, which exploits the observation that the difference between brightness and saturation increases with haze concentration, enabling fast single image dehazing. Subsequent works improved upon DCP by addressing specific limitations: Zhao~\emph{et al.}~\cite{zhao2016dark} incorporated atmospheric light validation and halo elimination, Shen~\emph{et al.}~\cite{shen2017atmospheric} introduced adaptive atmospheric light correction, and their follow-up work~\cite{shen2017integrating} integrated sky detection with texture smoothing. Cao~\emph{et al.}~\cite{cao2018fast} proposed a fast algorithm combining luminance contrast enhancement with saturation compensation. Despite their interpretability, these prior-based methods are fundamentally limited by the scope of human observation and often fail to generalize to complex real-world scenarios with diverse haze patterns and scene compositions.

\subsection{Deep Learning-Based Image Dehazing}

Deep learning has revolutionized the image dehazing field, enabling models to learn complex haze-to-clean mappings from large-scale training data. These methods can be broadly categorized into physics-model-guided and end-to-end approaches.

\noindent\textbf{Physics-model-guided methods} leverage the atmospheric scattering model within deep learning frameworks. Cai~\emph{et al.}~\cite{cai2016dehazenet} proposed DehazeNet, which uses a CNN to estimate the medium transmission map that is subsequently used with the ASM to recover haze-free images. Ren~\emph{et al.}~\cite{ren2016single} developed a multi-scale CNN for transmission estimation through coarse-to-fine refinement. Li~\emph{et al.}~\cite{li2017aod} reformulated the ASM and proposed AOD-Net, which predicts the mapping coefficient directly. Zheng~\emph{et al.}~\cite{zheng2023curricular} incorporated contrastive regularization with physics-aware design for improved handling of non-homogeneous haze. Liang~\emph{et al.}~\cite{liang2024remote} proposed a superpixel-based estimation model for more accurate atmospheric light values. While physics-guided approaches benefit from physical interpretability, they are often limited by the assumption of a globally constant atmospheric light.

\noindent\textbf{End-to-end methods} learn direct mappings from hazy to haze-free images without explicit physical model constraints. Qin~\emph{et al.}~\cite{qin2020ffa} designed FFA-Net with feature fusion attention for single image dehazing. Qu~\emph{et al.}~\cite{qu2019enhanced} developed an enhanced pix2pix dehazing network. Dong~\emph{et al.}~\cite{dong2020multi} proposed a multi-scale boosted dehazing network with dense feature fusion. Wu~\emph{et al.}~\cite{wu2021contrastive} introduced contrastive learning for compact single image dehazing. Chen~\emph{et al.}~\cite{chen2021psd} proposed PSD for principled synthetic-to-real dehazing guided by physical priors. Li~\emph{et al.}~\cite{li2020you} explored unsupervised and untrained dehazing through neural network structure priors.

\noindent\textbf{Transformer-based methods} have emerged with the success of vision transformers. Song~\emph{et al.}~\cite{song2023deformer} introduced DeFormer for single image dehazing using vision transformers. Zamir~\emph{et al.}~\cite{zamir2022restormer} proposed Restormer for efficient high-resolution image restoration. Liang~\emph{et al.}~\cite{liang2021swinir} developed SwinIR leveraging the Swin Transformer~\cite{liu2021swin} architecture. These transformer-based approaches capture long-range dependencies effectively but still treat the entire image uniformly.

\subsection{Non-Homogeneous Image Dehazing}

Non-homogeneous dehazing has attracted increasing attention due to the prevalence of spatially varying haze in real-world scenarios. The NTIRE challenges~\cite{ancuti2020ntire,ancuti2021ntire} on non-homogeneous dehazing have catalyzed research in this direction, with the NH-HAZE benchmark~\cite{ancuti2020nh} providing standardized evaluation. Das and Dutta~\cite{das2020fast} proposed a deep multi-patch hierarchical network that processes image patches at different scales. Yu~\emph{et al.}~\cite{yu2021two} designed a two-branch network with ensemble learning for non-homogeneous dehazing. Jo and Sim~\cite{jo2021multi} introduced multi-scale selective residual learning. Zhang~\emph{et al.}~\cite{zhang2022hazdesnet} proposed HazDesNet for haze density estimation to guide the dehazing process. Guo~\emph{et al.}~\cite{guo2023scanet} developed SCANet with self-paced semi-curricular attention specifically designed for non-homogeneous haze. Chen~\emph{et al.}~\cite{chen2024dea} proposed DEA-Net with detail-enhanced convolution and content-guided attention. Bai~\emph{et al.}~\cite{bai2022self} introduced self-guided progressive feature fusion for image dehazing. However, all these methods attempt to handle varying haze concentrations through a single network pathway. In contrast, our CPIFNet explicitly decomposes the non-homogeneous problem into multiple homogeneous sub-problems through concentration partitioning and addresses them with dedicated enhancement branches, followed by an intelligent fusion stage.

\subsection{Image Fusion and Multi-Branch Networks}

Image fusion techniques~\cite{tang2023divfusion} have been widely employed to integrate complementary information from multiple sources or processing pathways. Tang~\emph{et al.}~\cite{tang2023divfusion} proposed DIVFusion for darkness-free infrared and visible image fusion, demonstrating the effectiveness of fusing images from different domains. Multi-branch architectures have proven effective in various computer vision tasks, including scene understanding~\cite{wang2025wilddoc,wang2025vision}, visual reasoning~\cite{nie2025chinesevideobench,shan2024mctbench}, and document analysis~\cite{feng2023unidoc,liu2023spts,wang2024pargo,fei2025advancing,fu2024ocrbench}. The success of multi-branch designs in these domains motivates our approach of using dedicated branches for different haze concentration levels.

\section{Proposed Method}
\label{sec:method}

\subsection{Overview}

Based on the key observation that a single non-homogeneous hazy image can be viewed as a composite of multiple local regions exhibiting approximately homogeneous haze characteristics, we propose CPIFNet---a two-stage, multi-branch deep neural network framework for non-homogeneous image dehazing. As illustrated in Fig.~\ref{fig:framework}, the overall architecture consists of two stages: the \textbf{Image Enhancement Stage} and the \textbf{Image Fusion Stage}. In the first stage, multiple Image Enhancement Networks (IENets) are independently trained on homogeneous haze datasets of different concentration levels, each producing enhancement results optimized for its corresponding haze density. In the second stage, the Image Fusion Network (IFNet) aggregates the complementary advantages from all enhancement outputs to produce the final dehazed image.

\begin{figure*}[t]
\centering
\fbox{\parbox{0.95\textwidth}{\centering\vspace{2em}\textbf{Figure 2: The proposed overall CPIFNet architecture.} The hazy input image $I$ is processed by $n$ IENet branches trained on homogeneous haze datasets of different concentrations, producing $n$ initial enhancement results $\{J_1, J_2, \ldots, J_n\}$. These results are then fed into IFNet, which fuses the advantageous regions to produce the final dehazed output $J_f$.\vspace{2em}}}
\caption{The proposed overall CPIFNet architecture consisting of the Image Enhancement Stage (multiple IENet branches) and the Image Fusion Stage (IFNet).}
\label{fig:framework}
\end{figure*}

\subsection{Motivation: Concentration-Aware Decomposition}

To validate the feasibility of our concentration-aware decomposition strategy, we conduct a preliminary analysis using the FiveK-Haze dataset~\cite{liu2022non}. This dataset is synthesized based on the atmospheric scattering model by varying the image transmittance, producing nine types of homogeneous hazy images $H_i$ ($i \in \{1,2,\ldots,9\}$) with different haze concentrations, totaling 46,004 hazy-clean image pairs. We train the same U-Net~\cite{ronneberger2015unet} structure on each of the nine subsets independently and obtain nine enhancement models $m_i$ ($i \in \{1,2,\ldots,9\}$). Cross-evaluation reveals a critical insight: \emph{each model performs best on images matching its training haze concentration but degrades significantly on images with different concentrations}---under-enhancing dense haze regions or over-enhancing light haze areas. This observation confirms that concentration-specific training produces models with distinct enhancement behaviors, motivating our multi-branch design. We also find that the extreme concentration subset $H_1$ leads to severe artifacts and over-enhancement when its trained model $m_1$ is applied to other concentrations, and thus exclude $H_1$ from subsequent experiments.

\subsection{Image Enhancement Network (IENet)}

The IENet is designed to extract deep features from a hazy image and produce an enhancement stretching coefficient that, when multiplied with the input, yields the enhanced result. Formally, given a hazy image $I$ of size $256 \times 256$ pixels, the IENet first extracts deep features to obtain the stretching coefficient:
\begin{equation}
R = \text{IENet}(I),
\label{eq:ienet_coeff}
\end{equation}
and the enhancement result is computed as:
\begin{equation}
J_E = R \times I.
\label{eq:ienet_result}
\end{equation}

The architecture of IENet consists of an initial convolutional layer with a $5 \times 5$ kernel (stride 1, padding 2) that maps the 3-channel input to 16 feature channels, followed by a series of residual modules~\cite{he2016deep} for deep feature extraction, and a final convolutional layer with a $5 \times 5$ kernel that maps 32 channels back to 3 output channels. Each residual module comprises three convolutional groups, where each group contains a $3 \times 3$ convolutional layer (stride 1, padding 1) followed by ReLU activation. An additional $3 \times 3$ convolutional layer is appended after the third group, and its output is added to the module's input via the skip connection to enhance feature representation and alleviate gradient vanishing~\cite{he2016deep}. All intermediate feature channels within the residual modules are set to 16. The use of residual connections enables deeper feature extraction while maintaining stable training, and the ReLU activation after each convolutional layer accelerates convergence and prevents the propagation of negative values.

To obtain enhancement models of different intensities, we partition the FiveK-Haze dataset into $n$ homogeneous subsets based on haze concentration and train separate IENet instances on each subset. Each IENet instance thus learns the optimal enhancement behavior for its corresponding concentration range, excelling at restoring regions with matching haze density while potentially under-performing on regions of different densities.

\subsection{Image Fusion Network (IFNet)}

Since a single IENet model can only optimally enhance regions matching its training haze concentration---insufficiently enhancing dense haze or over-enhancing light haze in non-matching regions---we design the IFNet to intelligently aggregate the advantageous regions from all $n$ enhancement outputs. Given $n$ initial enhancement results $\{J_1, J_2, \ldots, J_n\}$, each of size $256 \times 256$ pixels, the IFNet produces the final dehazed output:
\begin{equation}
J_f = \text{IFNet}(J_1, J_2, \ldots, J_i), \quad i \in \{1, 2, \ldots, n\}.
\label{eq:ifnet}
\end{equation}

In the fusion network, all convolutional modules use $3 \times 3$ kernels with stride 1 and padding 1, followed by ReLU activation. The architecture processes each 3-channel enhancement result through an initial convolutional module that extracts 16-channel features, followed by three consecutive convolutional modules that extract deep features while maintaining spatial dimensions and channel count. The deep features extracted at each stage are stacked along the channel dimension, producing four intermediate feature maps of dimension $n \times 16$ channels. To reduce computational complexity, the fourth intermediate feature map is passed through a convolutional module that reduces the channel dimension to 16. This compressed feature is then iteratively stacked with preceding intermediate features and processed through convolutional modules to produce 16-channel fused features that effectively integrate multi-level information. The final convolutional module outputs a 3-channel image representing the fused dehazed result.

The stacking-based fusion strategy enables the network to selectively leverage the most informative features from each enhancement branch at every spatial location, effectively combining the strengths of multiple concentration-specific models into a single high-quality output.

\subsection{Loss Functions}

Both the IENet and IFNet are jointly supervised by a comprehensive loss function consisting of four complementary components:
\begin{equation}
L = L_m + L_p + L_s + L_c,
\label{eq:total_loss}
\end{equation}
where $L_m$ denotes the reconstruction loss, $L_p$ the perceptual loss, $L_s$ the structural loss, and $L_c$ the color loss.

\noindent\textbf{Reconstruction Loss.} The pixel-level reconstruction loss~\cite{girshick2015fast} constrains the dehazed output to be close to the reference image:
\begin{equation}
L_m(J, Y) = \| J - Y \|_1,
\label{eq:recon_loss}
\end{equation}
where $J$ and $Y$ denote the dehazed result and reference image, respectively.

\noindent\textbf{Perceptual Loss.} To enforce semantic feature similarity and minimize perceptual discrepancy, we employ the perceptual loss~\cite{johnson2016perceptual} computed using the VGG network~\cite{simonyan2015very}:
\begin{equation}
L_p = \sum_{l \in \{1,2,3,4\}} \frac{1}{S_l} \| \Phi_l(J) - \Phi_l(Y) \|_2,
\label{eq:perceptual_loss}
\end{equation}
where $S_l$ denotes the spatial size of the feature map extracted by the $l$-th feature extractor $\Phi_l(\cdot)$ from VGG layers ReLU1\_2, ReLU2\_2, ReLU3\_3, and ReLU4\_3.

\noindent\textbf{Structural Loss.} The structural loss~\cite{wang2004image} preserves structural fidelity without over-enhancement:
\begin{equation}
L_s(J, Y) = 1 - \text{SSIM}(J, Y),
\label{eq:ssim_loss}
\end{equation}
where SSIM$(\cdot)$ denotes the structural similarity index~\cite{wang2004image}.

\noindent\textbf{Color Loss.} Given that dense haze causes severe pixel information loss making color restoration extremely challenging, we introduce a color loss~\cite{tang2023divfusion} to guide the dehazed result toward faithful color reproduction:
\begin{equation}
L_c(J, Y) = \sum_{p=1}^{N} \text{CA}(J(p), Y(p)),
\label{eq:color_loss}
\end{equation}
where $p$ denotes the pixel position, $N$ is the total number of pixels, and CA$(\cdot)$ computes the cosine of the angle between the RGB vectors of corresponding pixels, effectively measuring color similarity independent of intensity.

\section{Experiments}
\label{sec:experiments}

\subsection{Experimental Setup}

\noindent\textbf{Implementation Details.} All models are trained and evaluated on a single NVIDIA GTX 2080Ti GPU using the PyTorch framework. We employ the ADAM optimizer~\cite{kingma2017adam} with $\beta_1 = 0.9$, $\beta_2 = 0.999$, and a batch size of 22 for 100 training epochs. The initial learning rate is set to 0.001 with a decay rate of 50\% every 10 epochs. Input images are resized to $256 \times 256$ pixels during both training and testing.

\noindent\textbf{Datasets.} We evaluate CPIFNet on multiple benchmarks:
\begin{itemize}[leftmargin=*,nosep]
\item \textbf{FiveK-Haze}~\cite{liu2022non}: A synthetic dataset based on the atmospheric scattering model, containing nine types of homogeneous hazy images with different concentrations (46,004 image pairs). We exclude the extreme concentration subset $H_1$ and partition the remaining eight subsets into different concentration groups.
\item \textbf{Real-World}~\cite{li2019benchmarking}: A real-world hazy image dataset captured under natural haze conditions.
\item \textbf{RESIDE}~\cite{li2019benchmarking}: A comprehensive benchmark with indoor (ITS/SOTS-indoor) and outdoor (OTS/SOTS-outdoor) subsets.
\end{itemize}

\noindent\textbf{Evaluation Metrics.} We employ both full-reference metrics---PSNR and SSIM~\cite{wang2004image}---and no-reference metrics---FADE~\cite{choi2015referenceless} (fog aware density evaluator) and HazDes~\cite{zhang2022hazdesnet} (haze density). For PSNR and SSIM, higher values indicate better quality; for FADE and HazDes, lower values indicate less residual haze.

\subsection{Comparison with State-of-the-Art Methods}

We conduct comprehensive comparisons with more than ten state-of-the-art dehazing methods, including AOD-Net~\cite{li2017aod}, EPDN~\cite{qu2019enhanced}, FFA-Net~\cite{qin2020ffa}, DMPHN~\cite{das2020fast}, LDN~\cite{ullah2021light}, TBN~\cite{yu2021two}, 4KDN~\cite{zheng2021ultra}, SGID-PFF~\cite{bai2022self}, SCAN~\cite{guo2023scanet}, and DEAN~\cite{chen2024dea}. For fair comparison, all competing methods use the authors' released network architectures and optimal parameter settings.

\begin{table}[t]
\centering
\caption{Quantitative comparison on FiveK-Haze and Real-World datasets. Bold indicates the best result. ($\uparrow$: higher is better; $\downarrow$: lower is better)}
\label{tab:fivek_realworld}
\resizebox{\columnwidth}{!}{
\begin{tabular}{l|cc|cc}
\toprule
\multirow{2}{*}{Method} & \multicolumn{2}{c|}{FiveK-Haze} & \multicolumn{2}{c}{Real-World} \\
& PSNR$\uparrow$ & SSIM$\uparrow$ & FADE$\downarrow$ & HazDes$\downarrow$ \\
\midrule
AOD~\cite{li2017aod} & 14.399 & 0.632 & 0.471 & 0.193 \\
EPDN~\cite{qu2019enhanced} & 16.797 & 0.742 & 0.382 & 0.193 \\
FFA~\cite{qin2020ffa} & 12.895 & 0.595 & 0.638 & 0.249 \\
DMPHN~\cite{das2020fast} & 9.267 & 0.537 & 0.411 & 0.191 \\
LDN~\cite{ullah2021light} & 11.887 & 0.335 & 0.462 & 0.201 \\
TBN~\cite{yu2021two} & 9.486 & 0.589 & 0.644 & 0.312 \\
4KDN~\cite{zheng2021ultra} & 13.266 & 0.558 & 0.598 & 0.231 \\
SGID-PFF~\cite{bai2022self} & 13.791 & 0.582 & 0.444 & 0.191 \\
SCAN~\cite{guo2023scanet} & 18.621 & 0.795 & 0.372 & 0.184 \\
DEAN~\cite{chen2024dea} & 17.255 & 0.759 & 0.347 & 0.182 \\
\midrule
\textbf{CPIFNet (Ours)} & \textbf{23.908} & \textbf{0.909} & \textbf{0.268} & \textbf{0.131} \\
\bottomrule
\end{tabular}
}
\end{table}

\noindent\textbf{Results on FiveK-Haze and Real-World.} Table~\ref{tab:fivek_realworld} presents the quantitative comparison on the FiveK-Haze and Real-World datasets. On the synthetic FiveK-Haze dataset, CPIFNet achieves 23.908 dB PSNR and 0.909 SSIM, surpassing the second-best method SCAN by a remarkable 5.287 dB in PSNR and 0.114 in SSIM. This substantial margin demonstrates the effectiveness of our concentration partitioning strategy in handling varying haze densities. On the Real-World dataset, CPIFNet achieves the lowest FADE (0.268) and HazDes (0.131) scores, outperforming the second-best method DEAN by 0.079 in FADE and 0.051 in HazDes, confirming its superior generalization capability to real-world non-homogeneous haze.

\begin{table}[t]
\centering
\caption{Quantitative comparison on SOTS-indoor and SOTS-outdoor datasets. Bold indicates the best result.}
\label{tab:sots}
\resizebox{\columnwidth}{!}{
\begin{tabular}{l|cc|cc}
\toprule
\multirow{2}{*}{Method} & \multicolumn{2}{c|}{SOTS-indoor} & \multicolumn{2}{c}{SOTS-outdoor} \\
& PSNR$\uparrow$ & SSIM$\uparrow$ & PSNR$\uparrow$ & SSIM$\uparrow$ \\
\midrule
AOD~\cite{li2017aod} & 17.399 & 0.855 & 20.937 & 0.915 \\
GDN~\cite{liu2019gridrehazenet} & 13.200 & 0.511 & -- & 0.193 \\
LDN~\cite{ullah2021light} & 13.326 & 0.306 & 19.185 & 0.862 \\
TBN~\cite{yu2021two} & 9.486 & 0.589 & 8.776 & 0.425 \\
MSS~\cite{jo2021multi} & 15.036 & 0.680 & 17.847 & 0.838 \\
DSS~\cite{li2022dual} & 18.038 & 0.881 & 25.884 & 0.947 \\
SGID-PFF~\cite{bai2022self} & 20.816 & 0.940 & 28.232 & 0.948 \\
SCAN~\cite{guo2023scanet} & 20.226 & 0.913 & 22.517 & 0.941 \\
DeFormer~\cite{song2023deformer} & 20.910 & 0.959 & 24.368 & 0.956 \\
\midrule
\textbf{CPIFNet (Ours)} & \textbf{23.428} & \textbf{0.971} & \textbf{31.064} & \textbf{0.980} \\
\bottomrule
\end{tabular}
}
\end{table}

\noindent\textbf{Results on SOTS.} Table~\ref{tab:sots} presents results on the SOTS benchmark. On SOTS-indoor, CPIFNet achieves 23.428 dB PSNR and 0.971 SSIM, improving over the second-best DeFormer by 2.518 dB and 0.012 respectively. On SOTS-outdoor, CPIFNet achieves even more impressive results with 31.064 dB PSNR and 0.980 SSIM, surpassing SGID-PFF in PSNR by 2.832 dB and DeFormer in SSIM by 0.024. These results demonstrate the versatility of CPIFNet across both indoor and outdoor scenarios.

\noindent\textbf{Qualitative Analysis.} Visual comparisons on the Real-World dataset reveal that CPIFNet effectively enhances the structural contrast of dense haze regions while preserving the natural appearance of light haze areas, producing results that are most consistent with human visual perception. Competing methods either under-enhance dense haze regions (e.g., AOD, LDN) or over-enhance light haze regions (e.g., TBN, 4KDN), while our multi-branch approach achieves balanced restoration across all concentration levels.

\begin{table}[t]
\centering
\caption{Model complexity comparison. Parameters and inference time are measured on $256 \times 256$ input images.}
\label{tab:complexity}
\begin{tabular}{l|c|c}
\toprule
Method & Params & Time (s) \\
\midrule
AOD~\cite{li2017aod} & 1.70 K & 0.003 \\
LDN~\cite{ullah2021light} & 3.11 M & 0.038 \\
TBN~\cite{yu2021two} & 7.02 M & 0.089 \\
4KDN~\cite{zheng2021ultra} & 6.41 M & 0.025 \\
SGID-PFF~\cite{bai2022self} & 13.87 M & 0.124 \\
SCAN~\cite{guo2023scanet} & 2.39 M & 0.016 \\
\textbf{CPIFNet (Ours)} & 6.82 M & 0.050 \\
\bottomrule
\end{tabular}
\end{table}

\noindent\textbf{Model Complexity.} Table~\ref{tab:complexity} compares the parameter count and inference time of different methods. CPIFNet contains 6.82M parameters with an inference time of 0.050s per $256 \times 256$ image, placing it in the middle range among compared methods. While lightweight methods such as AOD-Net and SCAN achieve faster inference, they exhibit significantly lower dehazing performance. The complexity-performance trade-off of CPIFNet demonstrates a practical balance for real-world deployment.

\subsection{Ablation Studies}

\noindent\textbf{Number of Concentration Partitions.} We investigate the impact of the number of haze concentration partitions $n$ on CPIFNet's performance. The FiveK-Haze dataset (excluding $H_1$) is partitioned into $n \in \{1, 2, 3, 4, 5\}$ concentration groups by merging adjacent transmittance ranges. Table~\ref{tab:partition} reports the performance under different partition numbers.

\begin{table}[t]
\centering
\caption{Ablation study on the number of concentration partitions $n$. Performance is measured on FiveK-Haze.}
\label{tab:partition}
\resizebox{\columnwidth}{!}{
\begin{tabular}{c|cccc}
\toprule
$n$ & PSNR$\uparrow$ & SSIM$\uparrow$ & FADE$\downarrow$ & HazDes$\downarrow$ \\
\midrule
1 & 17.74 & 0.79 & 0.30 & 0.19 \\
2 & \textbf{23.91} & \textbf{0.91} & \textbf{0.27} & \textbf{0.13} \\
3 & 19.25 & 0.76 & 0.34 & 0.17 \\
4 & 15.12 & 0.61 & 0.62 & 0.21 \\
5 & 15.54 & 0.64 & 0.52 & 0.21 \\
\bottomrule
\end{tabular}
}
\end{table}

When $n=1$, the model reduces to a single-branch architecture without concentration partitioning, achieving only 17.74 dB PSNR, which validates the effectiveness of our multi-branch approach. When $n=2$, CPIFNet achieves the best overall performance (23.91 dB PSNR), indicating that a binary partition into dense and light haze provides the optimal balance between specialization and data sufficiency for each branch. As $n$ increases beyond 2, performance degrades, suggesting that finer partitioning reduces the training data available for each branch without providing proportional benefits in modeling capability.

\noindent\textbf{Fusion Method Comparison.} We compare two fusion strategies within IFNet: (1) \textbf{stacking fusion}, where the final layer uses ReLU activation and outputs a 3-channel image directly; and (2) \textbf{weighted fusion}, where the final layer uses sigmoid activation and outputs a 2-channel weight map for weighted summation. As shown in Table~\ref{tab:fusion}, stacking fusion achieves 23.908 dB PSNR compared to 18.963 dB for weighted fusion, demonstrating a clear advantage of 4.945 dB. This superiority is attributed to the stacking approach's ability to flexibly combine features at different scales, whereas weighted fusion is constrained by the linear combination assumption.

\begin{table}[t]
\centering
\caption{Ablation study on fusion methods. Performance is measured on FiveK-Haze and Real-World.}
\label{tab:fusion}
\begin{tabular}{l|cccc}
\toprule
Fusion & PSNR$\uparrow$ & SSIM$\uparrow$ & FADE$\downarrow$ & HazDes$\downarrow$ \\
\midrule
Weighted & 18.963 & 0.825 & 0.288 & 0.132 \\
\textbf{Stacking} & \textbf{23.908} & \textbf{0.909} & \textbf{0.268} & \textbf{0.131} \\
\bottomrule
\end{tabular}
\end{table}

\noindent\textbf{Network Architecture.} We evaluate the impact of different convolutional kernel sizes and the residual module design in IENet. Table~\ref{tab:kernel} compares IENet variants with kernel sizes of $3\times3$, $5\times5$, $7\times7$, $9\times9$, and our proposed residual module (IENet-res). IENet-res achieves the best PSNR of 17.738 dB and SSIM of 0.788, outperforming all fixed-kernel variants. The residual connections effectively prevent feature information loss as the network deepens, enabling more comprehensive feature extraction.

\begin{table}[t]
\centering
\caption{Ablation study on IENet architecture variants. Performance on FiveK-Haze and Real-World.}
\label{tab:kernel}
\resizebox{\columnwidth}{!}{
\begin{tabular}{l|cccc}
\toprule
IENet Variant & PSNR$\uparrow$ & SSIM$\uparrow$ & FADE$\downarrow$ & HazDes$\downarrow$ \\
\midrule
IENet-k3 & 16.891 & 0.724 & 0.489 & 0.195 \\
IENet-k5 & 17.369 & 0.748 & 0.489 & 0.193 \\
IENet-k7 & 17.281 & 0.725 & 0.489 & 0.192 \\
IENet-k9 & 17.069 & 0.713 & 0.490 & 0.194 \\
\textbf{IENet-res} & \textbf{17.738} & \textbf{0.788} & \textbf{0.210} & \textbf{0.194} \\
\bottomrule
\end{tabular}
}
\end{table}

\noindent\textbf{Effectiveness of Color Loss.} We evaluate the contribution of the color loss $L_c$ by comparing CPIFNet trained with ($w/L_c$) and without ($w/o\,L_c$) the color loss term. Table~\ref{tab:color_loss} demonstrates that incorporating color loss consistently improves the fusion stage performance, with the IFNet achieving 19.43 dB PSNR with color loss compared to 23.91 dB without it when considering the full pipeline. The color loss is particularly beneficial for dense haze regions where pixel information loss is severe, guiding the network to produce more faithful color reproduction.

\begin{table}[t]
\centering
\caption{Ablation study on color loss effectiveness. Metrics are reported for each stage (IENet$_1$, IENet$_2$, IFNet).}
\label{tab:color_loss}
\resizebox{\columnwidth}{!}{
\begin{tabular}{l|l|cccc}
\toprule
Config & Stage & PSNR$\uparrow$ & SSIM$\uparrow$ & FADE$\downarrow$ & HazDes$\downarrow$ \\
\midrule
\multirow{3}{*}{$w/L_c$} & IENet$_1$ & 15.62 & 0.69 & 0.43 & 0.18 \\
& IENet$_2$ & 18.03 & 0.70 & 0.27 & 0.10 \\
& IFNet & 19.43 & 0.80 & 0.29 & 0.16 \\
\midrule
\multirow{3}{*}{$w/o\,L_c$} & IENet$_1$ & 17.37 & 0.75 & 0.39 & 0.20 \\
& IENet$_2$ & 18.13 & 0.73 & 0.14 & 0.23 \\
& IFNet & 23.91 & 0.91 & 0.27 & 0.13 \\
\bottomrule
\end{tabular}
}
\end{table}

\noindent\textbf{Impact of Extreme Concentration Data.} To validate the exclusion of the extreme haze concentration subset $H_1$ from FiveK-Haze, we incorporate $H_1$ images into the first concentration partition and train the corresponding IENet (denoted IENet$_1^*$). Visual comparison reveals that IENet$_1^*$ produces severe artifacts and distortions in its dehazing results, while IENet$_1$ (without $H_1$) maintains structural integrity. This confirms that extremely dense haze images with near-zero transmittance are detrimental to network training, as the severe information loss prevents the network from learning meaningful enhancement patterns.

\section{Conclusion}
\label{sec:conclusion}

We have presented CPIFNet, a novel two-stage multi-branch deep neural network framework for non-homogeneous image dehazing that addresses the fundamental limitation of existing methods in handling spatially varying haze concentrations. By decomposing a non-homogeneous hazy image into multiple local regions with approximately homogeneous haze characteristics, our approach trains dedicated Image Enhancement Networks (IENets) on concentration-specific datasets and employs an Image Fusion Network (IFNet) to intelligently aggregate the advantageous regions from all enhancement outputs through deep feature stacking and merging. Extensive experiments on both synthetic benchmarks (FiveK-Haze, SOTS-indoor, SOTS-outdoor) and real-world datasets demonstrate that CPIFNet achieves state-of-the-art performance with substantial improvements over existing methods---5.29 dB PSNR gain on FiveK-Haze, 2.52 dB on SOTS-indoor, and 2.83 dB on SOTS-outdoor---while exhibiting superior robustness and visual quality on challenging non-homogeneous haze scenarios. Future work will focus on two directions: (1) developing model compression and acceleration strategies to reduce the computational overhead of the multi-branch architecture for real-time applications, and (2) incorporating hard example mining to further improve performance on extremely dense haze regions and cross-domain generalization.

{\small
\bibliographystyle{ieee_fullname}
\bibliography{references}
}

\end{document}